\documentclass[10pt,twocolumn]{ICCAS}

\usepackage{comment} 
\usepackage{amsmath,amssymb,amsfonts}
\usepackage{algorithm}
\usepackage{algpseudocode}
\usepackage{multirow}
\usepackage{fancyvrb}
\usepackage{pdflscape}
\usepackage{lscape}
\usepackage{array}
\usepackage[caption=false,font=normalsize,labelfont=sf,textfont=sf]{subfig}
\usepackage{textcomp}
\usepackage{stfloats}
\usepackage{url}
\usepackage{verbatim}
\usepackage{graphicx}
\usepackage{cite}
\usepackage{subcaption}
\usepackage{caption}

\usepackage{amssymb}
\usepackage{orcidlink}
\usepackage{algorithm}
\usepackage{algpseudocode}
\usepackage{booktabs}
\usepackage{tikz}
\usepackage{diagbox}
\usetikzlibrary{arrows.meta, positioning, fit, backgrounds, calc, shapes.geometric, patterns, decorations.pathreplacing}

\begin{document}

\title{OSDAG: Online Scheduling for Efficient Multi-Robot Collaboration}

\author{
    Thanh Nguyen Canh$^{1,2\dagger}$\,\orcidlink{0000-0001-6332-1002}, 
    Thang Tran Viet$^{2\dagger}$\,\orcidlink{0009-0003-4442-9271}, 
    Phuc Van Dinh$^{2}$, \\
    Xiem HoangVan$^{2*}$\,\orcidlink{0000-0002-7524-6529},  
    and Nak Young Chong$^{1,3*}$\,\orcidlink{0000-0001-5736-0769}
    \thanks{$^\dagger$Equal contribution, $^*$Corresponding author.}
    \thanks{$^1$School of Information Science, JAIST, Ishikawa, Japan. (\texttt{\{thanhnc, nakyoung\}@jaist.ac.jp})}
    \thanks{$^2$VNU University of Engineering and Technology, Hanoi, Vietnam. (\texttt{\{23020773, 23020756, xiemhoang\}@vnu.edu.vn})}
    \thanks{$^3$Department of Robotics, Hanyang University, Gyeonggi, Korea.}
}

\affils{\textsuperscript{1}School of Information Science, Japan Advanced Institute of Science and Technology, Nomi, 923-1211, Ishikawa, Japan. \\ (\texttt{\{thanhnc, nakyoung\}@jaist.ac.jp}) {\small$^\dagger$Equal contribution}\\
\textsuperscript{2}University of Engineering and Technology, Vietnam National University, 10000, Hanoi, Vietnam. \\ (\texttt{\{23020773, 23020756, xiemhoang\}@vnu.edu.vn}) {\small${}^{*}$ Corresponding author}\\
\textsuperscript{3}Department of Robotics, Hanyang University, Ansan, 15588, Gyeonggi, Korea. \\}


\abstract{
Coordinating heterogeneous multi-robot systems (MRS) for complex, long-horizon tasks requires both flexible high-level reasoning and efficient execution-time scheduling. Existing LLM-based approaches struggle to balance reasoning efficiency and execution flexibility. Flat sequential plans are efficient to generate but overlook parallel execution opportunities, while repeated LLM reasoning introduces high latency, and offline schedules may unnecessarily keep robots idle due to fixed execution orders. 
This paper presents OSDAG, a novel framework that resolves this trade-off by employing a Directed Acyclic Graph (DAG) as the central representation for multi-robot coordination, coupled with constraint-aware online scheduling. The LLM is typically invoked once as a semantic parser that decomposes a natural-language instruction into a dependency-annotated task graph encoding precedence relations, together with robot capability and resource-feasibility constraints. A lightweight online scheduler then dynamically dispatches dependency-ready tasks to their assigned robots as soon as they become idle, exposing available parallelism while preserving correctness. Experiments across five benchmark scenarios demonstrate that OSDAG achieves $5-15\times$ faster reasoning time than dialogue-based methods, reduces makespan by up to $38\%$ over sequential baselines, and maintains competitive success rates. Both simulation and real-world experiments on human-robot collaboration tasks validate the effectiveness and practicality of the proposed approach for efficient multi-robot coordination.
The website and resources are available at \url{http://thanhnguyencanh.github.io/LLM_DAG4MultiRobot}
}


\maketitle


\section{Introduction}
Multi-Robot Systems (MRS) enable heterogeneous robots with distinct capabilities to collaboratively accomplish complex tasks. Compared to single-robot systems, MRS offer superior task completion time, broader workspace coverage, and flexible utilization of individual expertise, making them essential for search and rescue, warehouse automation, and industrial manufacturing~\cite{li2025large}. Traditional MRS coordination relies on manually designed control logic and planning algorithms requiring explicit behavior specification and deep domain knowledge~\cite{gerkey2004formal, giordani2010distributed}. While ensuring accuracy and predictability, these methods lack flexibility—any task or environment change necessitates costly reprogramming. Furthermore, classical approaches cannot interpret natural language, creating a gap between human intent and robot execution.

Recent advances in Large Language Models (LLMs) such as Gemini 2.5 Pro~\cite{comanici2025gemini} and GPT-4~\cite{achiam2023gpt} have opened new directions for robot coordination. Systems like SayCan~\cite{ahn2022can} and ChatGPT-Prompts\cite{wake2023chatgpt} demonstrate that natural language commands can be translated into executable robot actions with strong zero-shot generalization, enabling robots to adapt to novel tasks without reprogramming. Several studies have integrated LLMs into MRS using both centralized and decentralized paradigms. Decentralized methods like RoCo~\cite{mandi2024roco} assign each robot an LLM agent for dialogue-based coordination, but the conversation history grows linearly with the number of robots, leading to higher latency than single-query approaches. Centralized methods reduce this overhead but typically generate flat action sequences without modeling inter-task dependencies. More critically, existing frameworks—including recent works that combine LLMs with no-task-dependence execution~\cite{wang2024dart}, planning~\cite{song2023llm}, or Linear Programming~\cite{obata2024lip}—employ offline scheduling pipelines in which the entire plan is computed before any robot moves. This can force robots to follow predetermined execution orders. Consequently, a robot that finishes its current task may remain idle while waiting for an unrelated branch, even when another task assigned to the same robot is already executable.

Recent works further combine LLMs with structured planners to improve long-horizon coordination. LaMMA-P~\cite{zhang2025lamma} integrates LLM reasoning with PDDL planning, REBEL~\cite{gupte2024rebel} enhances task allocation using experience-based rules, ELHPlan~\cite{ling2025elhplan} improves planning efficiency through intention-aware collaboration, and CoMuRoS~\cite{borate2025llm} introduces event-driven replanning for heterogeneous teams. Other approaches incorporate dependency-aware execution~\cite{wang2024dart}, symbolic planning~\cite{song2023llm,bai2024twostep}, or optimization-based scheduling~\cite{obata2024lip}. In particular, LiP-LLM~\cite{obata2024lip} combines dependency graphs with linear-programming-based task allocation, while DART-LLM~\cite{wang2024dart} models subtask dependencies for sequential and parallel execution. In contrast, OSDAG uses lightweight execution-time scheduling: robot-task assignments remain fixed after LLM decomposition, while dependency-ready tasks are continuously re-evaluated and dispatched to idle robots, without repeated LLM inference or online optimization.

To address these limitations, we propose OSDAG, a framework that integrates LLM-based reasoning with a Directed Acyclic Graph (DAG) representation and online scheduling. The LLM decomposes high-level instructions into dependency-annotated subtasks formalized as a DAG. Unlike offline methods, our online scheduler continuously monitors robot and task states, dispatching ready tasks as soon as their assigned robots become idle. This enables independent subtasks to execute in parallel without waiting for unrelated task branches to complete. Our main contributions can be summarized as follows:
\begin{itemize}
    \item A mechanism that combines LLM reasoning with DAG-based dependency modeling, where the LLM generates a directed graph capturing inter-task dependencies based on robot capabilities and environmental constraints.
    \item An online scheduling algorithm that dynamically dispatches dependency-ready tasks to idle robots from their pre-assigned task sets, enabling concurrent execution while reducing unnecessary waiting.
    \item Experiments conducted in both simulation and real-world scenarios demonstrate superior performance in terms of success rate, reasoning latency, and execution efficiency compared to baseline models.
\end{itemize}

\begin{figure*}[!ht]
    \centering
    \includegraphics[width=0.91\linewidth]{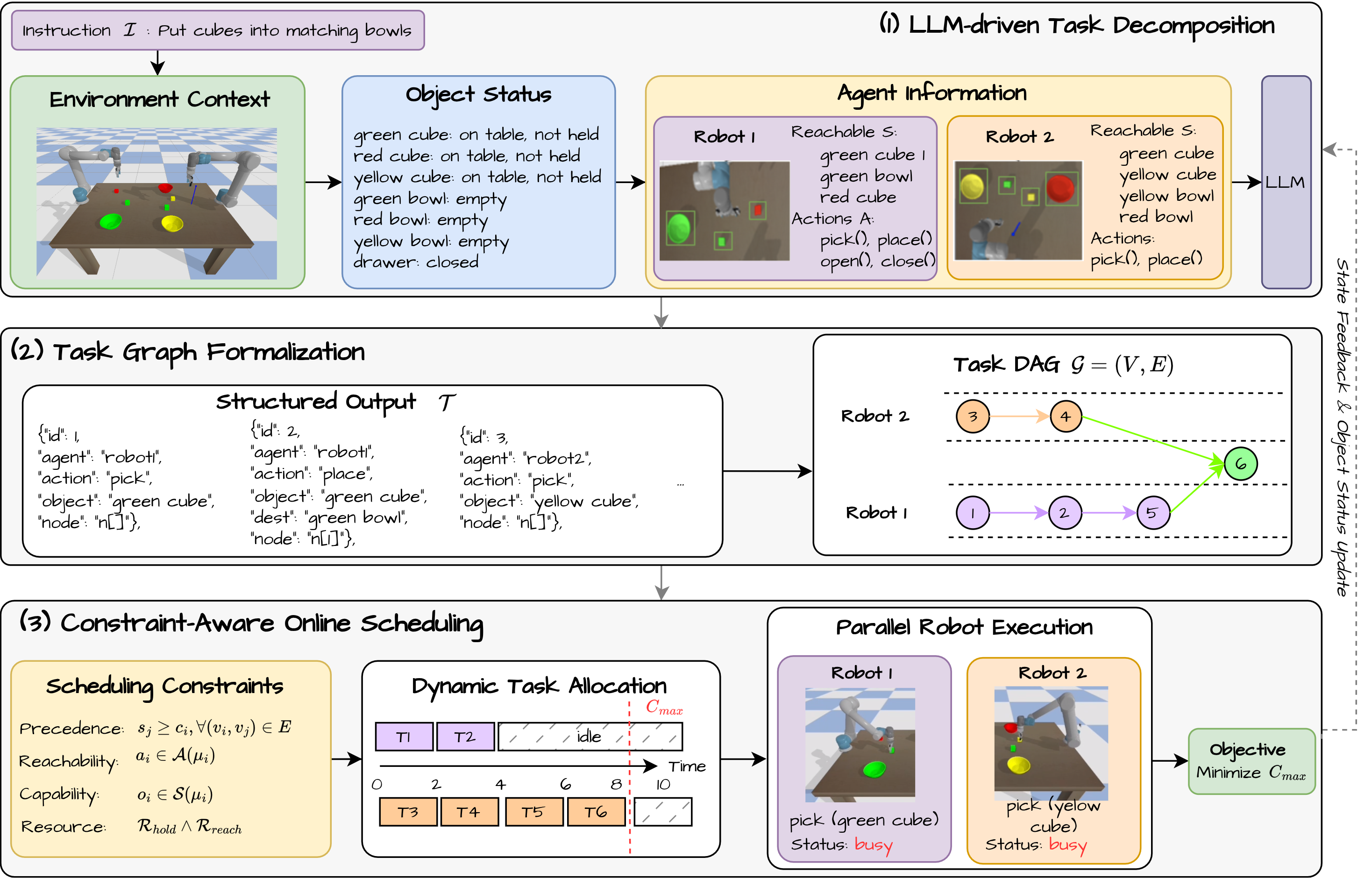}
    \caption{System overview of OSDAG. The framework processes a high-level natural language instruction through three stages: (1)~\textit{LLM-driven Task Decomposition} grounds the instruction into robot-assigned, dependency-annotated sub-tasks using environment context and capability descriptions; (2)~\textit{Task Graph Formalization} constructs a DAG encoding precedence and resource constraints; (3)~\textit{Constraint-Aware Online Scheduling} dynamically dispatches dependency-ready tasks to their assigned robots in real time, maximizing parallel execution while preserving task dependencies.}
    \label{fig:overview}
\end{figure*}

\section{Methodology}~\label{problemformulation}
OSDAG converts a natural-language instruction into a dependency-aware task graph and executes it using a lightweight online scheduler. Given an instruction ${\mathcal{I}}$, the LLM generates robot-assigned subtasks and their dependencies based on the current workspace. After validation, ready tasks are dispatched to idle robots while preserving precedence and feasibility constraints. Fig.~\ref{fig:overview} summarizes the overall pipeline.
\subsection{Problem Statement}
\label{subsec:problem_statement}

Consider a team of heterogeneous robots ${\mathcal{N}}=\{N_1,\ldots,N_N\}$ operating in a shared workspace containing objects ${\mathcal{O}}=\{o_1,\ldots,o_M\}$. Each robot $N_i$ has a reachable object set ${\mathcal{S}}(N_i) = \{o_j \in {\mathcal{O}} \mid d(N_i, o_j) \leq r_i\}$ and an executable action set ${\mathcal{A}}(N_i)$, such as \texttt{pick}, \texttt{place}, or \texttt{open}. The reachability between robot $N_i$ and object $o_j$ with their position $\mathbf{p}(N_i)$ and $\mathbf{p}(o_j)$ is determined by:
\begin{equation}
    d(N_i,o_j)=\|\mathbf{p}(N_i)-\mathbf{p}(o_j)\|_2,
\end{equation}
and encoded as a binary matrix $\mathbf{R} \in \{0,1\}^{N \times M}$:
\begin{equation}
    \mathbf{R}_{ij}={\mathbb{I}}[d(N_i,o_j)\leq r_i],
\end{equation}
where $r_i$ is the workspace radius of robot $N_i$ and ${\mathbb{I}}[\cdot]$ is the indicator function. Given a natural-language instruction ${\mathcal{I}}$, the objective is to generate a set of executable subtasks:
\begin{equation}
    {\mathcal{T}}=\{T_1,T_2,\ldots,T_m\},
\end{equation}
assign them to feasible robots, preserving their logical dependencies with the objective of reducing the overall makespan. Each subtask is represented as:
\begin{equation}
    T_i=(\mu_i,a_i,o_i,d_i,{\mathcal{D}}_i),
\end{equation}
where $\mu_i$ is the assigned robot, $a_i$ is the action, $o_i$ is the target object, $d_i$ is the destination or target state, and ${\mathcal{D}}_i \in \{id_1, \ldots, id_k\}$ denotes the prerequisite task set.

For each task $T_i$, let $s_i$, $p_i$, and $c_i=s_i+p_i$ denote its start time, execution duration, and completion time. The scheduling objective is
\begin{align}
\min \quad & C_{\max} \label{eq:objective}\\
\text{s.t.} \quad
& C_{\max} \geq c_i, && \forall i, \label{eq:makespan}\\
& s_j \geq c_i, && \forall (T_i,T_j)\in E, \label{eq:precedence}\\
& o_i \in {\mathcal{S}}(\mu_i), && \forall i, \label{eq:reachability}\\
& a_i \in {\mathcal{A}}(\mu_i), && \forall i. \label{eq:capability}
\end{align}
The constraints ensure that all precedence relations are respected and that each assigned robot can both reach the target object and execute the required action.

\begin{figure}[!t]
    \centering
    \includegraphics[width=\linewidth]{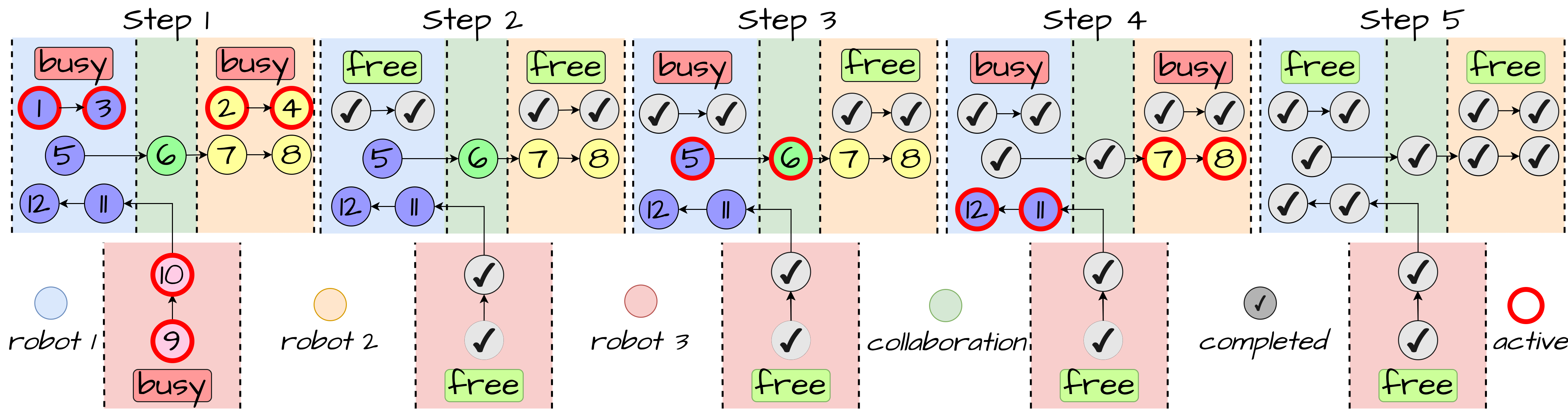}
    \caption{Illustration of the online scheduling process. Steps~(1)--(5) demonstrate dynamic task dispatching across three robots.}
    \label{fig:online_scheduling}
\end{figure}

\subsection{LLM-driven Task Decomposition}
\label{subsec:llm_decomposition}

The LLM is used as a high-level task reasoner rather than a low-level controller. Instead of repeatedly querying the LLM during execution, OSDAG typically invokes the LLM once to decompose the instruction into structured subtasks. The prompt is constructed as:
\begin{equation}
    {\mathcal{P}}=\Phi\left({\mathcal{I}},
    \{{\mathcal{S}}(N_i)\}_{i=1}^{N},
    \{{\mathcal{A}}(N_i)\}_{i=1}^{N},
    \boldsymbol{\sigma}\right),
\end{equation}
where $\boldsymbol{\sigma}$ denotes the current object states and workspace context. The prompt contains the user instruction, available robots, reachable objects, executable actions, and a strict output schema requiring each task to include its robot, action, object, destination, and dependencies.
The LLM generates a candidate task list:
\begin{equation}
    \hat{{\mathcal{T}}}=f_{\theta}({\mathcal{P}}),
\end{equation}
where $f_{\theta}$ denotes the LLM. Each generated task is validated before execution. A task $T_i$ is accepted only if
\begin{equation}
    {\mathcal{V}}(T_i)=
    \big(o_i\in{\mathcal{S}}(\mu_i)\big)
    \land
    \big(a_i\in{\mathcal{A}}(\mu_i)\big)
    \land
    \big({\mathcal{D}}_i \subseteq \{1,\ldots,m\}\big).
\end{equation}
This validation removes infeasible assignments, invalid actions, and incorrect dependency references. If the generated output violates the schema or feasibility constraints, the plan is rejected and regenerated with refined constraints.

\subsection{Task Graph Formalization}
\label{subsec:task_graph}

After validation, the structured task list is converted into a Directed Acyclic Graph ${\mathcal{G}}=(V,E)$. Each task $T_i$ corresponds to a vertex $v_i\in V$, and each prerequisite relation defines a directed edge:
\begin{equation}
    V=\{v_i \mid T_i\in{\mathcal{T}}\},
    \quad
    E=\{(v_j,v_i)\mid j\in{\mathcal{D}}_i\}.
\end{equation}
Thus, an edge $(v_j,v_i)$ indicates that task $T_j$ must be completed before task $T_i$ can start.
The graph ${\mathcal{G}}$ can also be represented by an adjacency matrix $\mathbf{A}\in\{0,1\}^{m\times m}$:
\begin{equation}
    \mathbf{A}_{ji}={\mathbb{I}}[(v_j,v_i)\in E].
\end{equation}
Each element of $\hat{\mathcal{T}}$ is returned as a JSON object containing an integer \texttt{id} and a \texttt{dependencies} list. A lightweight custom Python parser constructs one vertex per task and one edge per dependency, while acyclicity is verified by topological sort; cyclic graphs are rejected as logically inconsistent. The resulting DAG captures sequential and inter-robot dependencies, such as \texttt{pick}$\rightarrow$\texttt{place} and object handoffs, while exposing independent branches that can execute in parallel. Thus, a task can be dispatched as soon as its predecessors are completed, without waiting for unrelated branches.
\begin{figure}[]
\centering
\resizebox{\columnwidth}{!}{%
\begin{tikzpicture}[
    task/.style={minimum height=0.7cm, inner xsep=2pt, inner ysep=0pt,
                 outer sep=0pt, draw=black, thick, font=\bfseries, align=center},
    robot1task/.style={task, fill=blue!30},
    robot2task/.style={task, fill=orange!30},
    robot3task/.style={task, fill=purple!30},
    collabtask/.style={task, fill=green!30},
    idle/.style={task, fill=gray!10, draw=gray!50, pattern=north east lines, pattern color=gray!30},
    arrow/.style={-{Stealth[length=2mm]}, thick},
    timeline/.style={},
    agenttitle/.style={font=\bfseries},
]

\node[font=\large\bfseries] at (7, 2.9) {\Large Online Scheduling (Ours)};

\draw[thick, -{Stealth[length=3mm]}] (0, -1.0) -- (14.5, -1.0) node[right] {\Large Time (s)};
\foreach \x in {0, 2, 4, 6, 8, 10, 12, 14, 16, 18, 20} {
    \pgfmathsetmacro{\xpos}{\x * 0.7}
    \draw (\xpos, -1.1) -- (\xpos, -0.9);
    \node[below] at (\xpos, -1.1) {\Large \x};
}

\node[agenttitle, blue!70!black, anchor=east] at (-0.3, 1.5) {\Large Robot 1};
\node[agenttitle, orange!70!black, anchor=east] at (-0.3, 0.6) {\Large Robot 2};
\node[agenttitle, purple!70!black, anchor=east] at (-0.3, -0.3) {\Large Robot 3};

\node[robot1task, minimum width=2.1cm, anchor=west] (r1t1) at (0, 1.5) {\normalsize T1: pick};
\node[robot1task, minimum width=2.1cm, anchor=west] (r1t3) at (2.1, 1.5) {\normalsize T3: place};
\node[robot1task, minimum width=2.1cm, anchor=west] (r1t5) at (4.2, 1.5) {\normalsize T5: pick};
\draw[fill=green!30, draw=black, thick] (6.3, 1.15) rectangle (7.0, 1.85);
\node[font=\bfseries\large, inner sep=0pt] (r1t6a) at (6.65, 1.5) {T6};

\node[robot2task, minimum width=2.1cm, anchor=west] (r2t2) at (0, 0.6) {\normalsize T2: pick};
\node[robot2task, minimum width=2.1cm, anchor=west] (r2t4) at (2.1, 0.6) {\normalsize T4: place};
\node[idle, minimum width=2.8cm, minimum height=0.7cm, anchor=west] (r2idle1) at (4.2, 0.6) {\Large idle};
\node[robot2task, minimum width=1.4cm, minimum height=0.7cm, anchor=west] (r2t7) at (7, 0.6) {\Large T7};
\node[robot2task, minimum width=1.4cm, minimum height=0.7cm, anchor=west] (r2t8) at (8.4, 0.6) {\Large T8};

\node[robot3task, minimum width=2.1cm, anchor=west] (r3t9) at (0, -0.3) {\normalsize T9: pick};
\node[robot3task, minimum width=2.1cm, anchor=west] (r3t10) at (2.1, -0.3) {\normalsize T10: place};
\node[idle, minimum width=2.8cm, minimum height=0.7cm, anchor=west] (r3idle1) at (4.2, -0.3) {\Large idle};
\node[robot3task, minimum width=1.4cm, minimum height=0.7cm, anchor=west] (r3t11) at (7, -0.3) {\Large T11};
\node[robot3task, minimum width=1.4cm, minimum height=0.7cm, anchor=west] (r3t12) at (8.4, -0.3) {\Large T12};

\draw[thick, red, dashed] (9.8, -1.0) -- (9.8, 2.1);
\node[red, font=\bfseries] at (9.8, 2.3) {\Large$C_{max}=14s$};

\node[font=\large\bfseries] at (7, -2.2) {\Large Offline Scheduling (Baseline)};

\draw[thick, -{Stealth[length=3mm]}] (0, -5.5) -- (14.5, -5.5) node[right] {\Large Time (s)};
\foreach \x in {0, 2, 4, 6, 8, 10, 12, 14, 16, 18, 20} {
    \pgfmathsetmacro{\xpos}{\x * 0.7}
    \draw (\xpos, -5.6) -- (\xpos, -5.4);
    \node[below] at (\xpos, -5.6) {\Large \x};
}

\node[agenttitle, blue!70!black, anchor=east] at (-0.3, -3) {\Large Robot 1};
\node[agenttitle, orange!70!black, anchor=east] at (-0.3, -3.9) {\Large Robot 2};
\node[agenttitle, purple!70!black, anchor=east] at (-0.3, -4.8) {\Large Robot 3};

\node[robot1task, minimum width=2.1cm, minimum height=0.7cm, anchor=west] at (0, -3) {\Large T1};
\node[robot1task, minimum width=2.1cm, minimum height=0.7cm, anchor=west] at (2.1, -3) {\Large T3};
\node[robot1task, minimum width=2.1cm, minimum height=0.7cm, anchor=west] at (4.2, -3) {\Large T5};
\draw[fill=green!30, draw=black, thick] (6.3, -3.35) rectangle (7.0, -2.65);
\node[font=\bfseries\large, inner sep=0pt] at (6.65, -3) {T6};

\node[robot2task, minimum width=2.1cm, minimum height=0.7cm, anchor=west] at (0, -3.9) {\Large T2};
\node[robot2task, minimum width=2.1cm, minimum height=0.7cm, anchor=west] at (2.1, -3.9) {\Large T4};
\node[idle, minimum width=2.8cm, minimum height=0.7cm, anchor=west] at (4.2, -3.9) {\Large idle};
\node[robot2task, minimum width=1.4cm, minimum height=0.7cm, anchor=west] at (7, -3.9) {\Large T7};
\node[robot2task, minimum width=1.4cm, minimum height=0.7cm, anchor=west] at (8.4, -3.9) {\Large T8};

\node[robot3task, minimum width=2.1cm, minimum height=0.7cm, anchor=west] at (0, -4.8) {\Large T9};
\node[robot3task, minimum width=2.1cm, minimum height=0.7cm, anchor=west] at (2.1, -4.8) {\Large T10};
\node[idle, minimum width=5.6cm, minimum height=0.7cm, anchor=west] at (4.2, -4.8) {\Large idle};
\node[robot3task, minimum width=1.4cm, minimum height=0.7cm, anchor=west] at (9.8, -4.8) {\Large T11};
\node[robot3task, minimum width=1.4cm, minimum height=0.7cm, anchor=west] at (11.2, -4.8) {\Large T12};

\draw[thick, red, dashed] (12.6, -5.4) -- (12.6, -2.7);
\node[red, font=\bfseries, anchor=west] at (12.85, -2.75) {\Large $C_{max}=18s$};

\draw[thick, green!50!black, -{Stealth[length=3mm]}] (14.6, -0.9) -- (14.6, 1.8);
\node[green!50!black, font=\bfseries, rotate=90, anchor=south] at (15.2, 0.9) {\Large 22\% faster};

\begin{scope}[shift={(-1, -5.2)}]
    \node[robot1task, minimum width=1cm, minimum height=0.4cm] at (0, -1.5) {};
    \node[right] at (0.6, -1.5) {\Large Robot 1 Task};
    
    \node[robot2task, minimum width=1cm, minimum height=0.4cm] at (4.6, -1.5) {};
    \node[right] at (5.2, -1.5) {\Large Robot 2 Task};
    
    \node[collabtask, minimum width=1cm, minimum height=0.4cm] at (9.2, -1.5) {};
    \node[right] at (9.8, -1.5) {\Large Collaboration};
    
    \node[idle, minimum width=1cm, minimum height=0.4cm] at (14, -1.5) {};
    \node[right] at (14.6, -1.5) {\Large Idle Time};
\end{scope}

\end{tikzpicture}
}
\caption{Comparison of online scheduling (ours) and offline scheduling (baseline) on the three-robot task graph of Fig.~\ref{fig:online_scheduling}
\label{fig:timeline_compare}}
\end{figure}
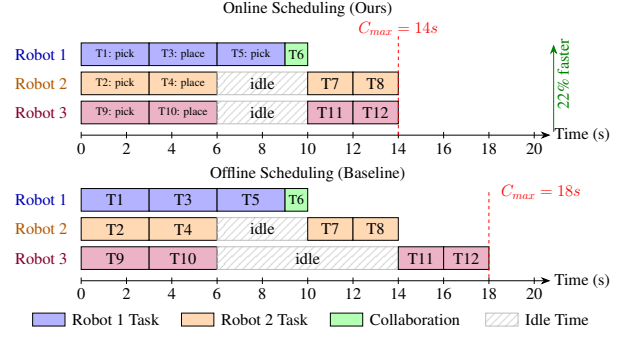

\subsection{Constraint-Aware Online Scheduling}
\label{subsec:online_scheduling}
Given the task graph ${\mathcal{G}}=(V,E)$, OSDAG performs online scheduling during execution. The scheduler maintains the completed-task set ${\mathcal{C}}(t)$, the running-task set ${\mathcal{B}}(t)$, and the state of each robot, either \textit{Free} or \textit{Busy}. For a free robot $N_i$, the executable task set at time $t$ is defined as:
\begin{equation}
\begin{split}
{\mathcal{E}}_i(t)&=\{T_j \mid \; \mu_j=N_i,\;
T_j\notin {\mathcal{C}}(t)\cup{\mathcal{B}}(t),\\
&\; \mathrm{Pred}(T_j)\subseteq{\mathcal{C}}(t),\;
{\mathcal{R}}(N_i,T_j,t)=\mathrm{true}\},
\end{split}
\end{equation}
where $\mathrm{Pred}(T_j)$ denotes the predecessor set of $T_j$. The resource feasibility term is defined as:
\begin{equation}
{\mathcal{R}}(N_i,T_j,t)=
{\mathcal{R}}_{\mathrm{hold}}(N_i,T_j,t)
\land
{\mathcal{R}}_{\mathrm{reach}}(N_i,T_j),
\end{equation}

where ${\mathcal{R}}_{\mathrm{hold}}$ checks manipulation-state constraints, such as whether the robot is holding the required object, and ${\mathcal{R}}_{\mathrm{reach}}$ checks whether the target object is reachable. Robot--task assignments remain fixed after LLM decomposition; the online scheduler only determines execution order and timing. Whenever robot $N_i$ becomes free, it re-evaluates the unfinished tasks assigned to $N_i$ and dispatches any task whose predecessors and runtime feasibility constraints are satisfied. Thus, a robot can proceed with another ready task without waiting for unrelated branches.

When multiple tasks are executable for the same robot, the earliest indexed task is selected:
\begin{equation}
    T_i^{*}(t)=
    \operatorname*{arg\,min}_{T_j\in{\mathcal{E}}_i(t)} j.
\end{equation}
After each task completion, the scheduler updates the completed-task set and re-evaluates newly executable tasks. This online procedure enables independent DAG branches to proceed in parallel while preserving dependency and feasibility constraints. Alg.~\ref{alg:online_scheduling} summarizes the procedure, while Fig.~\ref{fig:online_scheduling} and Fig.~\ref{fig:timeline_compare} illustrate the scheduling process and its comparison with fixed-order execution. This earliest-index rule is used only as a lightweight tie-breaker and does not guarantee a globally optimal schedule. More advanced priority- or optimization-based policies may further reduce makespan and are left for future work.

\begin{algorithm}[!t]
\caption{Constraint-Aware Online Scheduling}
\label{alg:online_scheduling}
\begin{algorithmic}[1]
\Require Task graph ${\mathcal{G}}=(V,E)$, robots ${\mathcal{N}}$
\Ensure Makespan $C_{\max}$
\State Initialize ${\mathcal{C}}\leftarrow\emptyset$, ${\mathcal{B}}\leftarrow\emptyset$
\State Set all robots to \textit{Free}
\While{${\mathcal{C}}\neq {\mathcal{T}}$}
    \For{each robot $N_i\in{\mathcal{N}}$}
        \If{$N_i$ is \textit{Free}}
            \State Compute executable task set ${\mathcal{E}}_i(t)$
            \If{${\mathcal{E}}_i(t)\neq\emptyset$}
                \State $T^*\leftarrow\operatorname*{arg\,min}_{T_j\in{\mathcal{E}}_i(t)} j$
                \State Dispatch $T^*$ to $N_i$
                \State Add $T^*$ to ${\mathcal{B}}$
            \EndIf
        \EndIf
    \EndFor
    \For{each completed task $T_j$}
        \State Move $T_j$ from ${\mathcal{B}}$ to ${\mathcal{C}}$
        \State Set the corresponding robot to \textit{Free}
    \EndFor
    \State $t\leftarrow t+\Delta t$
\EndWhile
\State \Return $C_{\max}=\max_i c_i$
\end{algorithmic}
\end{algorithm}

\begin{figure*}[!t]
    \centering
    \begin{minipage}[b]{0.15\textwidth}
        \centering
        \includegraphics[width=\textwidth]{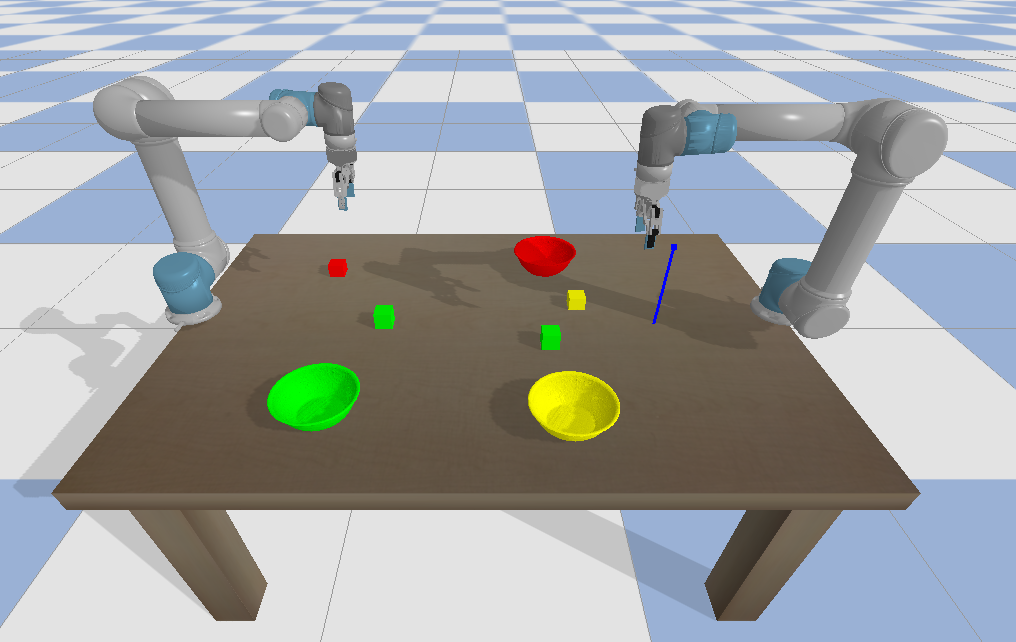}
        \caption*{(a) Task 1}
    \end{minipage}
    \hfill
    \begin{minipage}[b]{0.146\textwidth}
        \centering
        \includegraphics[width=\textwidth]{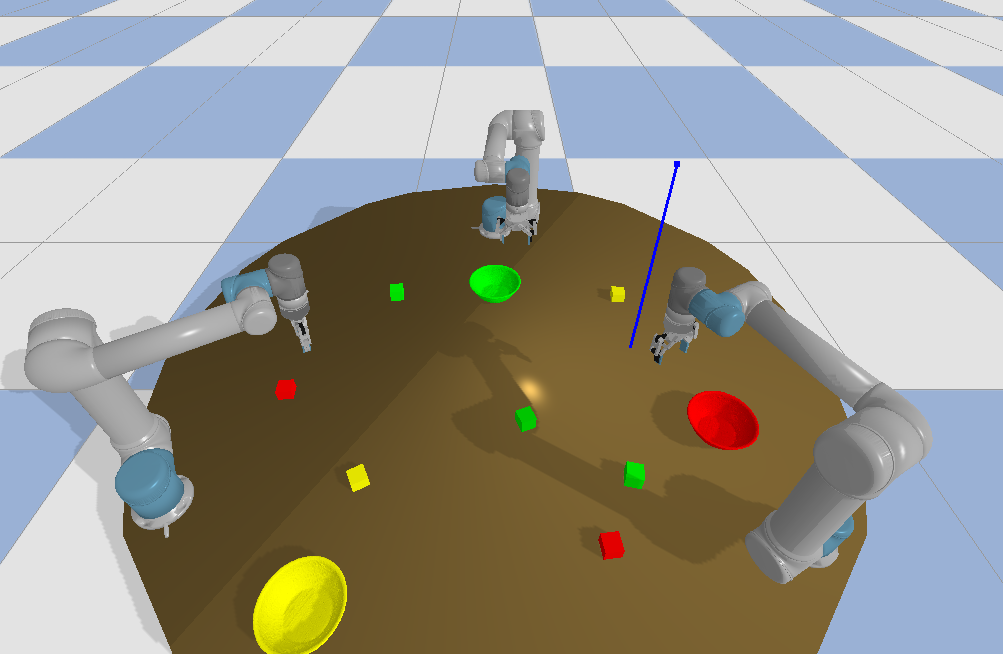}
        \caption*{(b) Task 2}
    \end{minipage}
    \hfill
    \begin{minipage}[b]{0.164\textwidth}
        \centering
        \includegraphics[width=\textwidth]{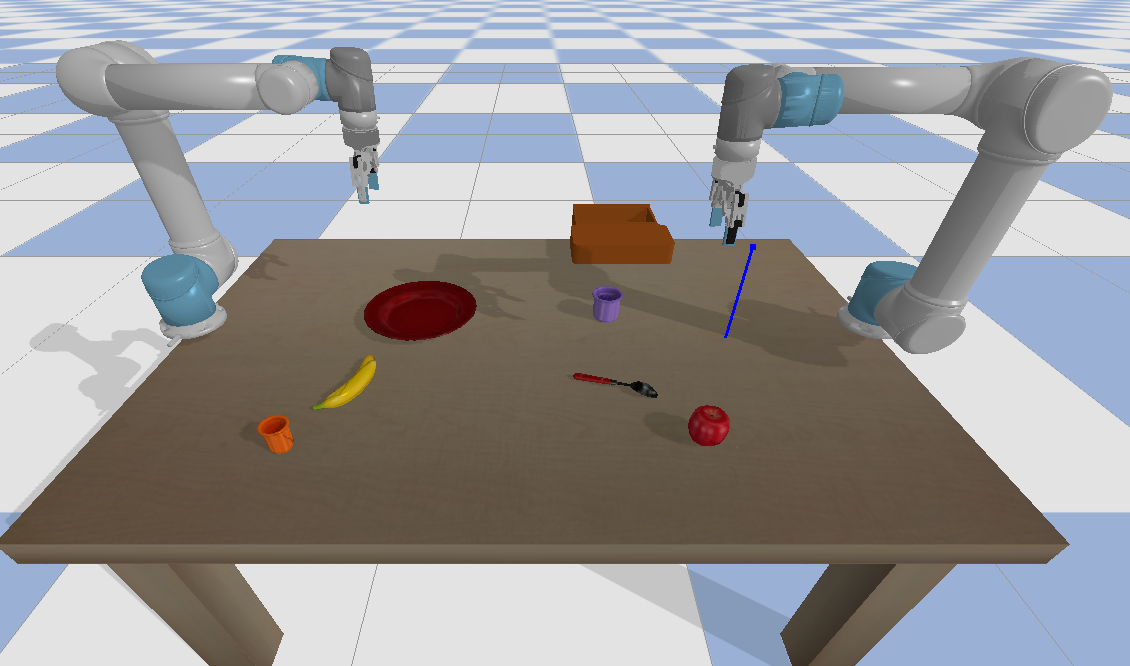}
        \caption*{(c) Task 3}
    \end{minipage}
    \hfill
    \begin{minipage}[b]{0.18\textwidth}
        \centering
        \includegraphics[width=\textwidth]{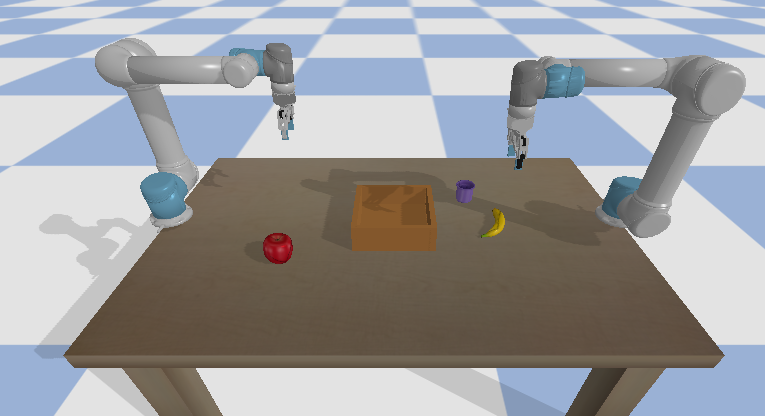}
        \caption*{(d) Task 4}
    \end{minipage}
    \hfill
    \begin{minipage}[b]{0.184\textwidth}
        \centering
        \includegraphics[width=\textwidth]{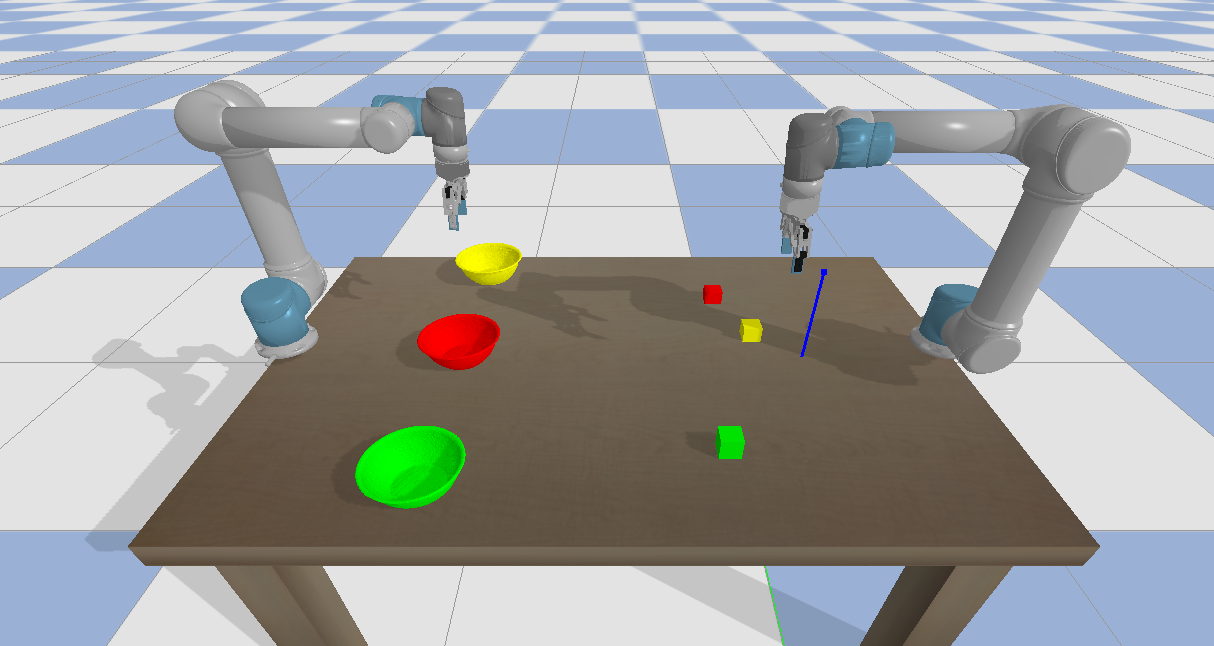}
        \caption*{(e) Task 5}
    \end{minipage}
    \caption{Simulation environments for the five evaluation tasks.}
    \label{fig:evaluation_tasks}
\end{figure*}

\section{Experimental Results}
\label{experiment}
\subsection{Experimental Setup}
We evaluate our system in both simulation and real-world environments. Simulation experiments are conducted using the PyBullet physics engine with custom tabletop manipulation environments. We employ Gemini Flash as the LLM backbone to validate planning efficiency and reasoning performance. To comprehensively assess performance, we design five evaluation scenarios covering different task complexities and spatial configurations. \textbf{Task 1} - Place cubes into matching bowls with 2 robots; \textbf{Task 2} - Place cubes into matching bowls with 3 robots to evaluate parallel execution capability. \textbf{Task 3} - Clean table to test implicit dependency reasoning (\textit{e.g.}, opening a drawer before placing items); \textbf{Task 4} - Sort objects in order: apple $\rightarrow$ banana $\rightarrow$ cup to assess sequential dependency handling. \textbf{Task 5} - Place cubes into matching bowls to evaluate inter-robot collaboration with object handoffs. 

\textbf{Evaluation Metrics}.
We evaluate performance using four complementary metrics: Planning Success Rate (SR, $\%$), Reasoning Time (RT, seconds), Makespan (MS, seconds), and Success-weighted Step Efficiency (SSE)~\cite{obata2024lip}.

\textbf{Baselines}.
We compare against three LLM-based multi-robot planning baselines: RoCo~\cite{mandi2024roco}, a dialogue-based decentralized framework; TwoStep~\cite{bai2024twostep}, a hierarchical LLM-PDDL planner; and ChatGPT-Prompts~\cite{wake2023chatgpt}, a direct flat-sequence generation method.All methods use the same task inputs, Gemini Flash backbone, generation settings, and hardware, with method-specific prompt templates.

\subsection{Simulation Results}
Table~\ref{tab:results} reports performance across all five tasks, averaged over
10 trials. OSDAG achieves substantially lower reasoning time than dialogue-based methods and reduces makespan in dependency-heavy tasks, while maintaining competitive success rates.

\begin{table*}[t]
\scriptsize
\caption{\textbf{Performance comparison across all tasks.} Best results are in \textbf{bold}. ``--'' indicates execution failure.}
\label{tab:results}
\centering
\setlength{\tabcolsep}{4pt}
\begin{tabular}{l|cccc|cccc|cccc|cccc|cccc}
\toprule
\multirow{2}{*}{\textbf{Method}}& \multicolumn{4}{c|}{\textbf{Task 1}} & \multicolumn{4}{c|}{\textbf{Task 2}} & \multicolumn{4}{c|}{\textbf{Task 3}} & \multicolumn{4}{c|}{\textbf{Task 4}} & \multicolumn{4}{c}{\textbf{Task 5}} \\
 & SR & SSE & RT & MS & SR & SSE & RT & MS & SR & SSE & RT & MS & SR & SSE & RT & MS & SR & SSE & RT & MS \\
\hline
RoCo~\cite{mandi2024roco} & 100 & 1.0 & 108.5 & 19.1 & 90 & 0.88 & 206.3 & \textbf{21.9} & 0 & 0 & 134.7 & -- & 100 & 1.0 & 64.1 & 22.7 & 100 & 1.0 & 94.8 & 41.6 \\ 
TwoStep~\cite{bai2024twostep} & 90 & 0.9 & 15.2 & 19.1 & 70 & 0.66 & 14.3 & 22.0 & 0 & 0 & 14.8 & -- & 0 & 0 & 14.7 & -- & 90 & 1.0 & 15.0 & 41.6 \\ 
ChatGPT-Prompts~\cite{wake2023chatgpt} & 100 & 0.5 & \textbf{8.2} & 33.9 & 90 & 0.36 & \textbf{8.9} & 35.9 & \textbf{90} & 0.51 & \textbf{8.3} & 48.3 & 100 & 1.0 & \textbf{10.3} & 22.7 & 100 & 1.0 & \textbf{9.2} & 41.6 \\ \hline
\textbf{Ours} & \textbf{100} & \textbf{1.0} & 9.1 & \textbf{19.1} & \textbf{90} & \textbf{0.88} & 14.0 & 22.5 & 80 & \textbf{0.8} & 13.4 & \textbf{26.1} & \textbf{100} & \textbf{1.0} & 12.8 & \textbf{15.2} & \textbf{100} & \textbf{1.0} & 10.5 & \textbf{26.0} \\ 
\hline
\end{tabular}
\end{table*}

\textbf{Reasoning Time.}
OSDAG requires 9.1–14.0 s per task, yielding a 5–15× speedup over RoCo. In Task 2, RoCo requires 206.3 s compared with 14.0 s for OSDAG due to its dialogue-based coordination. ChatGPT-Prompts achieves slightly lower RT through single-call flat-sequence generation, but its weaker dependency modeling results in lower SSE and substantially higher makespan.

\textbf{Success Rate.} Our system achieves 80\% SR on Task~3, a complex scenario involving implicit dependencies requiring LLM reasoning (e.g., opening a drawer before placing items). By embedding current object states $\boldsymbol{\sigma}$ into the prompt, OSDAG
surfaces this hidden precondition in most trials, outperforming both RoCo (0\%)
and TwoStep (0\%), which were not provided with the same explicit object-state grounding and failed to recover the hidden precondition in our experiments.

\textbf{Makespan.} Our system reduces makespan by 33\% in Task~4 (15.2 vs 22.7) and 38\% in Task~5 (26.0s vs 41.6s) compared to baselines. By fully utilizing idle time through online scheduling, robots avoid waiting for others---Robot~2 can retrieve a bowl immediately while Robot~1 is placing, eliminating idle periods prevalent in offline methods. This improvement comes from removing unnecessary synchronization between independent branches rather than from changing robot-task assignments.

\subsection{Ablation Study}

To isolate each component's contribution, we conduct ablation studies with: \textbf{w/o Context Grounding} - Removes object reachability, action descriptions, and status embeddings from the LLM prompt; \textbf{w/o Graph}: Generates sub-tasks without dependency annotations, yielding a flat sequence; \textbf{w/o Online Sched.}: uses a fixed per-robot task order. If the next task is blocked by an unmet dependency, the robot waits instead of executing another ready task in its task set. Table~\ref{tab:ablation} presents the results.
\begin{table*}[t]
\scriptsize
\caption{\textbf{Ablation study results.} We evaluate each component's contribution by systematic removal. Removing dependency graphs causes complete failure (SR$=$0).}
\label{tab:ablation}
\centering
\setlength{\tabcolsep}{4pt}
\begin{tabular}{l|cccc|cccc|cccc|cccc|cccc}
\toprule
\multirow{2}{*}{\textbf{Variant}} & \multicolumn{4}{c|}{\textbf{Task 1}} & \multicolumn{4}{c|}{\textbf{Task 2}} & \multicolumn{4}{c|}{\textbf{Task 3}} & \multicolumn{4}{c|}{\textbf{Task 4}} & \multicolumn{4}{c}{\textbf{Task 5}} \\
 & SR & SSE & RT & MS & SR & SSE & RT & MS & SR & SSE & RT & MS & SR & SSE & RT & MS & SR & SSE & RT & MS \\
\hline
w/o Context Grounding & 90 & 0.9 & 16.5 & 19.1 & 70 & 0.68 & 28.3 & 22.5 & 70 & 0.7 & 18.2 & 26.1 & 100 & 1.0 & 15.7 & 15.2 & 100 & 1.0 & 12.3 & 26.0 \\
w/o Graph & 0 & -- & \textbf{8.5} & -- & 0 & -- & \textbf{9.2} & -- & 0 & -- & \textbf{8.3} & -- & 0 & -- & \textbf{10.1} & -- & 0 & -- & 10.5 & -- \\
w/o Online Schedule & 100 & 1.0 & 9.1 & 19.1 & 90 & 0.88 & 14.0 & 22.5 & 80 & 0.8 & 13.4 & 26.1 & 100 & 1.0 & 12.8 & 22.7 & 100 & 1.0 & 10.5 & 41.6 \\
\hline
\textbf{Full (Ours)} & \textbf{100} & \textbf{1.0} & 9.1 & \textbf{19.1} & \textbf{90} & \textbf{0.88} & 14.0 & \textbf{22.5} & \textbf{80} & \textbf{0.8} & 13.4 & \textbf{26.1} & \textbf{100} & \textbf{1.0} & 12.8 & \textbf{15.2} & \textbf{100} & \textbf{1.0} & \textbf{10.5} & \textbf{26.0} \\
\hline
\end{tabular}
\end{table*}

\textbf{Effect of Context Grounding.} Removing grounded object, action, and state context increases reasoning time (16.5s vs 9.1s in Task~1) and reduces success rate (70\% vs 90\% in Task~2). Without grounded context, the LLM occasionally hallucinates invalid actions or incorrect robot assignments.

\textbf{Effect of Dependency Graph.} Without dependency annotations, all tasks fail (SR $= 0$) due to resource conflicts and lack of coordination. Robots attempt concurrent manipulation of the same object or violate precedence constraints, causing execution failures.

\textbf{Effect of Online Scheduling.} Disabling online scheduling increases makespan significantly despite correct plans---Task~4 rises from 15.2s to 22.7s (+49\%), and Task~5 from 26.0s to 41.6s (+60\%). This confirms that online task dispatching is essential for efficient parallel execution.

\subsection{Real-World Experiments}
To demonstrate practical applicability beyond simulation, we deploy our framework in a human-robot collaboration (HRC) scenario. The robotic agent is a UF850 manipulator with an Intel RealSense D435 RGB-D camera and a parallel gripper, while the human receives the scheduled actions in real time via a text display (\textit{e.g.}, ``Human: Task 5 -- pick(green\_cube)''). We employ YOLOv11-seg for object detection and instance segmentation, followed by depth alignment to obtain 3D object poses; the resulting object categories, positions, and dimensions are embedded into the LLM prompt for task decomposition.

\begin{figure*}
    \centering
    \includegraphics[width=0.8\linewidth]{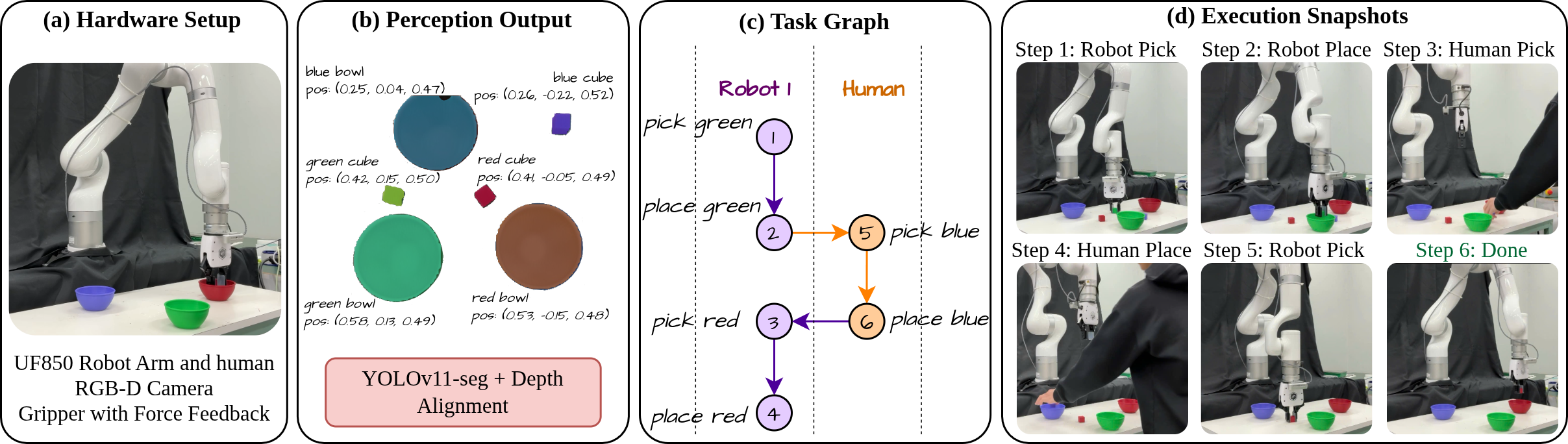}
    \caption{Qualitative results of real-world UF850 manipulation.}
    \label{fig:realworld_qualitative}
\end{figure*}

We evaluate a collaborative cube-sorting task where the robot and human work in parallel to place colored cubes into matching bowls. The human is modeled as a capability-constrained executor executing only dispatched tasks, so the experiment tests coordination across heterogeneous execution interfaces rather than human--robot equivalence. Fig.~\ref{fig:realworld_qualitative}(c) shows the generated task graph and Fig.~\ref{fig:realworld_qualitative}(d) shows execution snapshots at key timestamps, demonstrating that OSDAG effectively coordinates executors with different execution interfaces in real-world HRC scenarios.

\section{Conclusion}~\label{conclu}
We presented OSDAG, a novel framework for multi-robot coordination that integrates the high-level reasoning capabilities of Large Language Models with DAG-based task representation and real-time online scheduling. By embedding robot reachability, manipulation capabilities, and object states into structured prompts, the LLM generates dependency-aware subtasks that are scheduled online for execution by their pre-assigned robots. This design strictly preserves task dependencies while enabling parallel execution of independent subtasks, achieving 5–15$\times$ faster reasoning time than dialog-based methods and up to a 38\% reduction in makespan compared to offline scheduling approaches. However, our framework relies on the LLM to generate dependency edges, and reasoning errors may increase as environmental complexity and robot count increase. The current task selection policy may not yield globally optimal schedules. Additionally, the system lacks replanning mechanisms—execution failures can cascade without dynamic recovery. 

\section*{Acknowledgment}
This work was supported by JST SPRING, Japan Grant Number JPMJSP2102.

\bibliographystyle{IEEEtran}
\bibliography{ref}  

\newpage
\end{document}